\documentclass[sigconf,nonacm]{acmart}
\AtBeginDocument{%
  }

\usepackage{algorithm}
\usepackage{listings}
\usepackage{tabularx}
\usepackage{xcolor}
\usepackage{fancyvrb}
\usepackage{pgfplots}
\usepackage{booktabs, multirow}
\usepackage{tikz}
\usepackage{tcolorbox}
\usepackage{graphicx}
\usetikzlibrary{shapes.geometric, arrows.meta, positioning, fit, backgrounds, calc}

\pgfplotsset{compat=1.18}

\definecolor{baseline}{RGB}{149, 165, 166}      
\definecolor{cot}{RGB}{52, 152, 219}           
\definecolor{fewshot}{RGB}{231, 76, 60}        
\definecolor{positive}{RGB}{46, 204, 113}        
\definecolor{negative}{RGB}{231, 76, 60}         
\definecolor{mixed}{RGB}{155, 89, 182}           
\definecolor{contrastive}{RGB}{52, 152, 219}     
\definecolor{typematch}{RGB}{231, 76, 60}        
\definecolor{guideline}{RGB}{243, 156, 18}       
\definecolor{noteretrieval}{RGB}{52, 152, 219} 
\definecolor{optimal}{RGB}{46, 204, 113}         
\definecolor{kmore}{RGB}{243, 156, 18}           

\begin{document}

\title{Med-HEAL: Analyzing and Mitigating Hallucinations in Medical LLMs with Hallucination-Aware In-Context Learning}
\author{Yiming Liao}
\orcid{0009-0001-6583-970X}
\email{yliao1@umbc.edu}
\affiliation{%
  \institution{University of Maryland, Baltimore County}
  \city{Baltimore}
  \state{Maryland}
  \country{USA}
}
\additionalaffiliation{%
  \institution{Concordia University Wisconsin}
  \city{Grafton}
  \state{Wisconsin}
  \country{USA}
}

\author{Zeno Franco}
\orcid{0000-0002-6533-4226}
\email{zfranco@mcw.edu}
\affiliation{%
  \institution{Medical College of Wisconsin}
  \city{Milwaukee}
  \state{Wisconsin}
  \country{USA}
}

\author{Jose Eduardo Lizarraga Mazaba}
\orcid{0000-0001-8126-4048}
\email{jlizarraga@mcw.edu}
\affiliation{%
  \institution{Medical College of Wisconsin}
  \city{Milwaukee}
  \state{Wisconsin}
  \country{USA}
}

\author{Keke Chen}
\orcid{0000-0002-9996-156X}
\email{kekechen@umbc.edu}
\affiliation{%
  \institution{University of Maryland, Baltimore County}
  \city{Baltimore}
  \state{Maryland}
  \country{USA}
}

\renewcommand{\shortauthors}{Liao et al.}


\begin{abstract}

Hallucinations in medical large language models (LLMs) pose serious risks for clinical decision support, particularly when models must reason over complex electronic health records (EHRs). However, existing benchmarks often lack a realistic clinical context and provide limited insight into how hallucinations can be mitigated in practice. We introduce Med-HEAL, a framework for systematically identifying, analyzing, and mitigating hallucinations in medical LLMs using clinically grounded data. Building on the EHRNoteQA benchmark derived from MIMIC-IV discharge summaries, we construct a hallucination dataset by evaluating BioMistral-7B on open-ended clinical question answering tasks. Model outputs are labeled through a dual evaluation pipeline that combines LLM-as-a-Judge assessment (GPT-4o) with human auditing by medical student reviewers, producing correctness judgments and annotations of reasoning errors via a custom web-based evaluation system.

We then leverage this dataset to investigate mitigation strategies: a self-critique pipeline, in which the test model reviews its own answers to detect potential errors and regenerates responses for flagged cases, and retrieval-augmented in-context learning (RA-ICL), which exposes the model to hallucinated and corrected examples. Experiments across five open-source LLMs—BioMistral, Llama-3.1, DeepSeek, Qwen2.5, and Qwen3, show that the self-critique strategy improves accuracy for three of five models ($p < 0.05$) without requiring parameter updates.

Med-HEAL provides both a reusable hallucination dataset and a practical framework for studying and mitigating hallucinations in medical LLMs, supporting safer deployment of AI systems in clinical environments. Our code and data are publicly available at \url{https://github.com/yimingliao-blad/med-heal.git}.
\end{abstract}


\maketitle

\section{Introduction}

The integration of large language models (LLMs) into healthcare has the potential to transform how clinicians interact with medical data. Modern healthcare systems generate vast amounts of information through electronic health records (EHRs), which contain rich but largely unstructured clinical narratives \cite{singhalLargeLanguageModels2023}. LLMs provide powerful capabilities for analyzing such data, enabling automated information extraction from clinical notes, summarization of patient histories, and assistance with clinical reasoning tasks \cite{agrawalLargeLanguageModels2022}. By reducing the burden of navigating complex documentation, these systems may help alleviate the growing workload and information overload faced by healthcare professionals.

Recent research has adapted general-purpose LLMs to the medical domain using domain-specific corpora \cite{singhalLargeLanguageModels2023}, instruction tuning \cite{agrawalLargeLanguageModels2022, luoBioGPTGenerativePretrained2022, wangHuatuoTuningLlama2023}, and retrieval-augmented generation techniques \cite{lewisRetrievalaugmentedGenerationKnowledgeintensive2020, zakkaAlmanacRetrievalAugmented2024}. The long-term vision is to deploy LLMs as on-demand clinical assistants that can analyze a patient’s medical history, answer clinicians’ questions, and support diagnostic decision-making.

Despite these promising capabilities, the deployment of LLMs in healthcare remains constrained by the problem of \emph{hallucinations}---the generation of incorrect, fabricated, or unsupported medical information. While hallucinations in general-purpose applications may cause minor inconvenience, errors in clinical contexts can lead to serious consequences. Detecting such errors is particularly challenging because evaluating clinical reasoning often requires domain expertise. Recent studies have highlighted the prevalence of hallucinations in medical LLMs and proposed benchmark datasets to measure these failures \cite{agarwal2024medhalu,kim2025medical,panditMedhalluComprehensiveBenchmark2025a}.

However, several important challenges remain.

\begin{itemize}
\item \textbf{Non-clinically-realistic hallucinations in existing medical hallucination benchmarks.}
Many current benchmarks separate clinical questions from the complex contexts in which healthcare decisions are made. Datasets such as MedHalu \cite{agarwal2024medhalu} and MedHallu \cite{panditMedhalluComprehensiveBenchmark2025a} draw questions from structured knowledge bases including PubMedQA \cite{jinPubmedqaDatasetBiomedical2019}, LiveQA \cite{ben2019question}, USMLE \cite{jin2021disease}, and MedicationQA \cite{abacha2019overview}. While these sources provide high-quality medical questions, they primarily test factual recall rather than the reasoning required to interpret patient records. Moreover, many benchmarks generate synthetic hallucination examples. For example, MedHallu \cite{panditMedhalluComprehensiveBenchmark2025a} iteratively refines incorrect answers using GPT-4 to produce adversarial distractors. Such artificially constructed errors may differ substantially from the spontaneous reasoning failures that occur when LLMs analyze real clinical narratives.

\item \textbf{Limitations of fine-tuning-based mitigation methods.}
Fine-tuning has been widely used to reduce hallucinations in medical LLMs. For instance, MedCaseReasoning collects diagnostic reasoning cases from clinical reports and performs supervised fine-tuning to reduce logical errors in model outputs \cite{wu2025medcasereasoning}. However, fine-tuning presents several practical challenges. First, medical knowledge evolves rapidly as treatment guidelines, drug approvals, and clinical protocols are updated, which can quickly render fine-tuned model parameters outdated. Second, repeated fine-tuning may introduce catastrophic forgetting \cite{goodfellow2013empirical,kirkpatrick2017overcoming} or degrade general reasoning ability, potentially creating new forms of hallucination \cite{gekhman2024does}.
\end{itemize}

\paragraph{\textbf{Scope of This Work.}}
To better understand and mitigate hallucinations in clinical reasoning tasks, we introduce \textbf{Med-HEAL}, a framework for collecting realistic hallucination examples and using them to improve model reliability at inference time.

First, we construct a clinically grounded hallucination dataset by leveraging the existing dataset EHRNoteQA, a benchmark built on the MIMIC-IV database that links clinical questions to multiple discharge summaries. This dataset reflects the complexity of real clinical reasoning tasks, where models must synthesize information across long and heterogeneous medical narratives. We utilize the error outputs of BioMistral-7B, a fine-tuned medical-domain LLM, and evaluate them with a dual-judge pipeline that combines LLM-as-a-Judge assessment with human auditing, enabling scalable labeling while maintaining clinical fidelity.

Second, we propose an inference-time mitigation framework centered on self-critique. In this pipeline, the model first generates an answer, then critiques its own response to detect potential reasoning errors, retrieves relevant demonstrations from the Med-HEAL dataset, regenerates the answer when errors are identified, and finally verdicts whether to accept the corrected answer only when it is better than the original answer. We also investigate plain retrieval-augmented in-context learning (RA-ICL) using relevant hallucination examples from the Med-HEAL dataset. Our experiments show that while RA-ICL alone provides limited benefits, the self-critique mechanism consistently improves performance over zero-shot answers across multiple models with statistical significance.

In summary, this work makes the following contributions:

\begin{itemize}
\item We introduce \textbf{Med-HEAL}, a framework for collecting clinically grounded hallucination examples from LLM interactions with real EHR discharge summaries.
\item We propose an inference-time mitigation pipeline based on self-critique, enabling models to detect and correct reasoning errors without parameter updates.
\item Through experiments across five open-source LLMs, we demonstrate that self-critique consistently improves clinical question answering accuracy on EHRNoteQA, outperforming retrieval-based ICL and chain-of-thoughts prompting.
\end{itemize}

\section{Methods}
We will start with an overview of the components of the Med-HEAL framework, then describe the key contributions for each component.

\subsection{Overview}
We introduce \textbf{Med-HEAL}, a framework for systematically identifying and mitigating hallucinations in medical large language models (LLMs). The framework contains two core components:

\textbf{(1)} Hallucination Data Collection Pipeline, which extracts clinically realistic hallucination cases from LLM outputs using a dual-judge evaluation system combining LLM-based evaluation and human auditing.

\textbf{(2)} Self-critique and retrieval-augmented In-Context Learning, which incorporates hallucinated reasoning pinpoint and corrected format examples into retrieval-augmented prompts to guide LLM responses.

\begin{figure*}[htbp]
  \centering
  \includegraphics[width=0.8\linewidth]{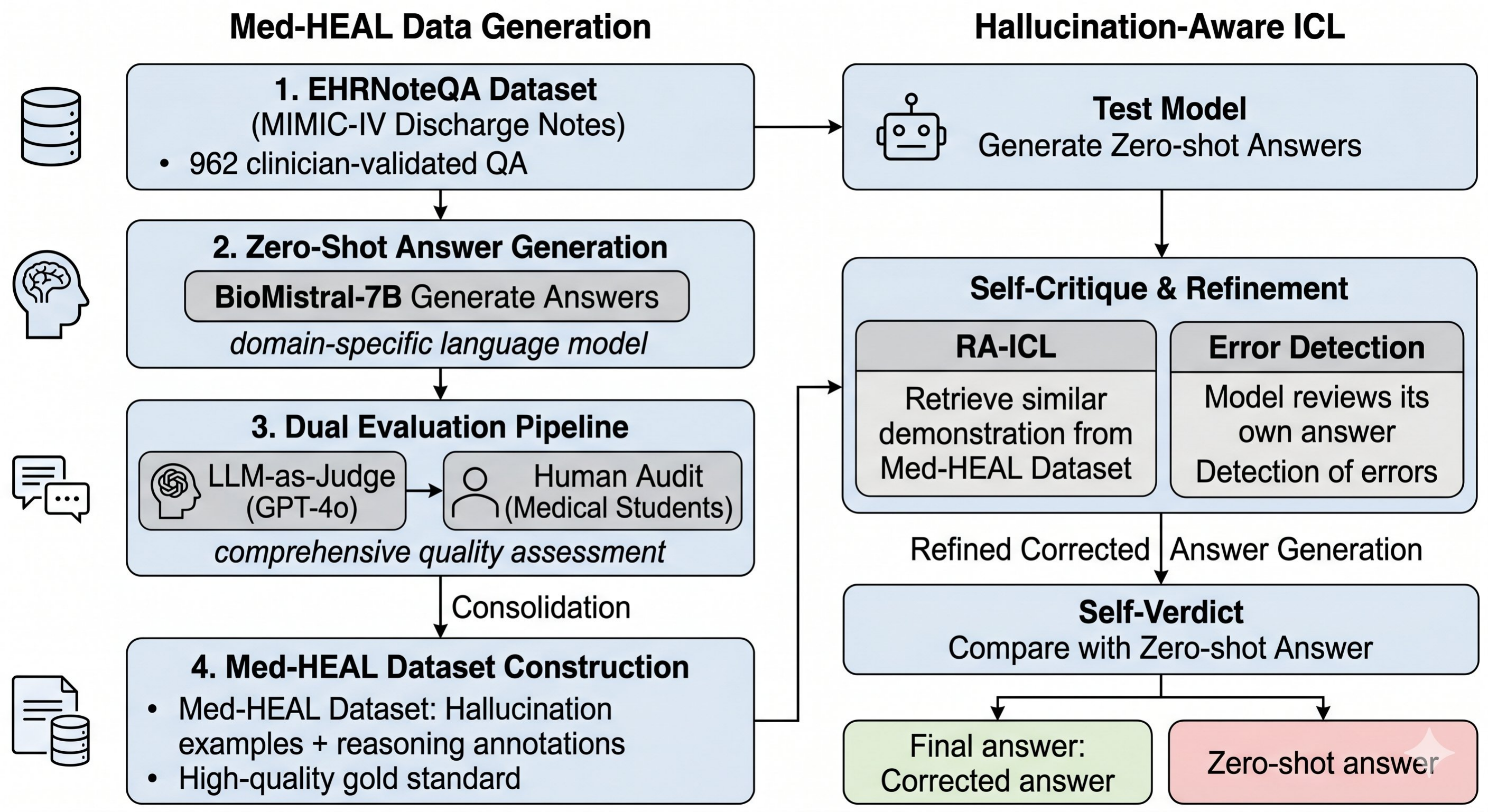}
  \caption{Overall Pipeline of Med-HEAL. Phase 1 (Left): LLM-generated labels are audited against human ground truth provided by medical student reviewers; these annotations are then consolidated to construct the hallucination dataset. Phase 2 (Right): The Self-Critique pipeline performs error detection and utilizes retrieval-augmented in-context learning (RA-ICL) with similar demonstrations to generate refined answers, which are then finalized through a self-verdict step.}
  \Description{A two-phase workflow diagram showing hallucination data collection from LLM answers and human audits, followed by self-critique, retrieval-augmented correction, and self-verdict for final answer generation.}
  \label{fig:overall-pipeline}
\end{figure*}

The overall pipeline is illustrated in Fig.~\ref{fig:overall-pipeline}. First, a medical LLM generates answers to EHR-based clinical questions. These outputs are evaluated by both an automated LLM judge and human reviewers to identify hallucinations and collect explanatory annotations. The resulting dataset is used to construct contrastive demonstrations for in-context learning on downstream models.

\subsection{Hallucination Data Collection Pipeline}

\paragraph{\textbf{Clinical QA Generation.}} The initial hallucination data used in the following study were generated by BioMistral-7B \cite{labrak2024biomistral} feeding on the original EHRNoteQA \cite{kweon2024ehrnoteqa} benchmark dataset. EHRNoteQA dataset is a novel benchmark that is specifically designed to evaluate LLMs in real-world clinical scenarios by answering clinicians’ questions regarding patient discharge summaries chosen from the Medical Information Mart for Intensive Care (MIMIC)-IV Clinical Database \cite{johnson2020mimic}, which selected 962 cases and manually created questions and answers in both open-ended and multiple-choice formats. EHRNoteQA covers a broad range of real-world clinical scenarios by categorizing queries into 10 diverse medical topics listed in Table~\ref{tab:ehrnoteqa-topics}. We select BioMistral because it is a representative, customized medical LLM that is prone to generating hallucinatory results, making it perfect for our hallucination data collection. BioMistral is based on the Mistral-7B-Instruct model, fine-tuned with a vast corpus of medical research papers from PubMed Central for complex medical question-answering and clinical reasoning tasks. 

We deliberately chose open-ended answers to generate hallucination data, even though EHRNoteQA supports both open-ended and multiple-choice formats. Multi-choice results include both a letter option (A-E) and a statement of reasoning, which increases the likelihood of contradictions and inconsistencies among human reviewers and LLM judges and could make the analysis of the results difficult and degrade the reliability of labeling in the next procedure. However, the open-ended format also increases the difficulty for human reviewers in examining BioMistral's output within the context of long discharge summary notes. For this reason, we have limited the current human validation set to 100 samples. 

\paragraph{\textbf{Dual-Judge Hallucination Identification}} Med-HEAL established a dual evaluation system that combines LLM-as-a-Judge and Human-as-a-Judge to balance the judgment cost and accuracy. We first use GPT-4o as the LLM-as-a-Judge to compare the ground-truth answer from EHRNoteQA and the summary from BioMistral to generate ``match/mismatch'' label. Then, we recruit medical students to evaluate the quality of LLM-as-a-Judge by manually labeling ``match/mismatch'' for 100 random samples from LLM judged results (63\% matched and 37\% unmatched, to match the distribution of LLM judge's results).

To improve the efficiency of human evaluation, we also developed a web application to assist the human reviewer. It provides the background discharge note(s), the ground-truth answer from the EHRNoteQA datasets, and BioMistral's answer. Among human reviewers, we use majority-vote to resolve the conflicts of disagreements, which is used as the gold standard for Human-as-a-Judge. To ensure fairness, we present the same information to the LLM judge and the human reviewers. The result can be found in Table~\ref{tab:biomistral_baseline}.


We further analyze how these hallucination examples are distributed across categories derived from prior work \cite{kim2025medical}. Table~\ref{tab:error_categories} shows these error categories. Because full human review of error rationales is resource-intensive, we use GPT-4o with category exemplars in the prompt to classify hallucination examples. Table~\ref{tab:biomistral_baseline} shows the resulting distribution, compared with reported error distributions in prior studies \cite{kim2025medical, kweon2024ehrnoteqa}.

\begin{table*}[ht]
\centering
\caption{Taxonomy of error types, and sample errors in MedHEAL.}
\label{tab:error_categories}
\small
\renewcommand{\arraystretch}{1.3}
\begin{tabular}{p{2.4cm} p{4.2cm} p{7.8cm}}
\toprule
\textbf{Error Type} & \textbf{Description} & \textbf{Example (Model Error Answer $\rightarrow$ Ground Truth)} \\
\midrule
1 Omission &
Missing documented information such as diagnoses, medications, or procedures. &
Q: \textit{What discharge medications were prescribed?} \\
& & \textcolor{red}{Model: ``Metoprolol and lisinopril.''} \\
& & \textcolor{teal}{Truth: ``Metoprolol, lisinopril, \textbf{warfarin, and atorvastatin}.''} \\[4pt]

2 Fabrication &
Fabricating facts not present in the discharge notes. &
Q: \textit{How was the fracture managed?} \\
& & \textcolor{red}{Model: ``Patient underwent \textbf{intramedullary nailing}.''} \\
& & \textcolor{teal}{Truth: ``Patient underwent \textbf{total hip replacement}.''} \\[4pt]

3 Reasoning Failure &
Misinterpreting clinical information (e.g., suspected vs.\ confirmed diagnosis). &
Q: \textit{What was the final diagnosis?} \\
& & \textcolor{red}{Model: ``\textbf{Pneumonia}'' (initial working diagnosis).} \\
& & \textcolor{teal}{Truth: ``\textbf{Congestive heart failure}'' (confirmed at discharge).} \\[4pt]

4 Specificity &
Providing vague answers when specific values are documented. &
Q: \textit{What were the patient's potassium levels?} \\
& & \textcolor{red}{Model: ``Potassium was \textbf{low}.''} \\
& & \textcolor{teal}{Truth: ``Potassium was \textbf{3.1\,mEq/L}.''} \\[4pt]

5 Context Confusion &
Conflating information across multiple discharge summaries. &
Q: \textit{What procedure was performed during the second admission?} \\
& & \textcolor{red}{Model: ``\textbf{CABG}'' (from 1st admission).} \\
& & \textcolor{teal}{Truth: ``\textbf{Valve replacement}'' (2nd admission).} \\[4pt]

6 Temporal Error &
Incorrect chronological ordering of clinical events. &
Q: \textit{Were antibiotics started before or after culture results?} \\
& & \textcolor{red}{Model: ``Started \textbf{after} cultures returned.''} \\
& & \textcolor{teal}{Truth: ``Empiric antibiotics initiated \textbf{two days prior}.''} \\
\bottomrule
\end{tabular}
\end{table*}

\paragraph{\textbf{Med-HEAL Dataset Construction.}} The Med-HEAL dataset is constructed by gathering BioMistral model outputs on the\linebreak EHRNoteQA benchmark. Specifically, each entry contains $r = (D, q, a, a^*, y, e)$, which include the EHRNoteQA context (discharge summary $D$), the clinical question $q$, the BioMistral-generated answer $a$, the ground truth answer $a^*$, the LLM judge's output $y \in \{0, 1\}$, where $y=1$ indicates a hallucination case, and the main error $e$ extracted from $a$ when $y=1$. Unlike synthetic hallucination benchmarks, Med-HEAL captures naturally occurring hallucinations from realistic EHR reasoning tasks, enabling the analysis of real-world failure modes.

To facilitate efficient retrieval of correct and hallucination examples, each component of the entry, i.e., $D$, $q$, $a$, $a^*$, and $e$, is vectorized for similarity-based retrieval. After testing a set of similarity-based retrieval measures, such as BM25 \cite{robertson2009probabilistic}, KATE \cite{wang2020minilm}, and GTR-T5-Base encoding \cite{ni2022large}, we adopt GTR-T5-base encoding, as it yielded the best retrieval performance for semantic similarity when applied to RA-ICL. 

\subsection{Hallucination-Aware In-Context Learning}
To mitigate clinical reasoning errors, we introduce \textbf{hallucination-aware in-context learning (ICL)}. This method utilizes documented hallucination instances from the Med-HEAL Dataset to guide the correction of detected errors, thereby reducing overall hallucination rates. The framework includes the following core components:

\paragraph{\textbf{Self-Critique.}}
We employ a self-critique and retrieval framework to select contrastive few-shot cases that guide the downstream model in correcting its own reasoning, thereby producing a hallucination-free response. 

The self-critique framework consists of two steps: error detection and correction. Given a test case $(D_t, q_t)$, the target model first generates a zero-shot answer $a_t$. In the first detection step, given $D_t$, $q_t$, and $a_t$, the target model will try to identify the most significant error, denoted by $e_t$. If the detection finds no error, we accept the zero-shot answer $a_t$ without any changes. Otherwise, we will proceed to the correction step. 

The error correction step uses the detected error $e_t$ to retrieve a contrastive few-shot case from the Med-HEAL lookup set. Given the current case $(D_t, q_t)$, we query hallucination records, i.e., $r_i = (D_i,\allowbreak q_i,\allowbreak a_i,\allowbreak a_i^*,\allowbreak y_i,\allowbreak e_i)$ with $y_i=1$, and score each candidate by discharge-summary similarity:

\[
\mathrm{sim}(D_t, D_i) = \cos(\phi(D_t), \phi(D_i)),
\]

Here, $\phi(\cdot)$ denotes GTR-T5 encoding, which converts the given discharge-note text into an embedding representation; the record with the highest similarity score is retrieved as the most relevant case.

\paragraph{\textbf{Self-Verdict.}}
Finally, a self-verdict mechanism using a prompt similar to the detection step in self-critique to detect potential errors in the refined answer. If the self-verdict indicates that the correction leads to more errors than the zero-shot answer $a_t$, the system rejects the revision and retains $a_t$. Otherwise, the revised response is accepted as the output of the whole \textbf{hallucination-aware in-context learning (ICL)} pipeline.

\subsection{Experiment Setup}

\textbf{Hardware and Software Environment.}
Our experiments were conducted on a local workstation equipped with an Intel Core i9-13900K processor (24 cores, 32 threads), 251 GB RAM, and a single NVIDIA GeForce RTX 4090 GPU (24 GB VRAM), running Ubuntu 24.04 LTS. Local models were served via the vLLM framework for efficient inference. All GPT-4o API calls for the dual-judge evaluation were made through the official OpenAI API. The runtime comparison is listed in Appendix \ref{sec:supp_runtime}.

\textbf{Baselines.} We compare hallucination-aware prompting with several baseline prompting strategies, including zero-shot prompting, positive retrieval (correct examples only), negative retrieval (hallucination examples with corrections), naive contrastive (one correct and one hallucination example) retrieval, and Chain-of-Thought prompting. Check Table~\ref{tab:icl_methods} for a detailed description of these methods. 

Unlike parameter fine-tuning approaches, the proposed method transfers hallucination knowledge entirely through prompt demonstrations. This allows the framework to be applied to multiple open-source LLMs without retraining. 

\textbf{ICL data split.} For the retrieval-augmented ICL (RA-ICL) experiments, we use 5-fold set. Each round, one fold is used as the retrieval example set and the other four folds are used for testing.

\textbf{Tested local medical LLMs} serve different roles in the pipeline. BioMistral-7B is used as the source model to collect hallucination cases. Qwen2.5-7B-Instruct is used as the pilot model for designing the transfer ICL architecture, and the finalized setup is then evaluated across the full model set. GPT-4o is used to judge answers from the tested models.

\textbf{In-context learning:} We evaluate Qwen2.5-7B \cite{yang2024qwen25technicalreport}, Qwen3-8B, LLaMA-3-8B \cite{dubey2024llama3herdmodels}, and DeepSeek-R1-Distill-LLaMA-8B \cite{deepseek_r1_distill_llama_8b}, served via vLLM (temperature=0.1, maxtokens=8,096), under 5-fold cross-validation on 962 questions from EHRNoteQA. Answer correctness is assessed by GPT-4o as binary (yes/no), comparing against the ground-truth option text.


All prompts follow the ChatML format with a system message and a user message. The shared user prompt contains \texttt{Discharge Summary} and \texttt{Question} fields.

The system message varies by condition; the base instruction tells the model to answer questions about discharge summaries as a medical expert.

For ICL conditions, demonstration examples and instructions are appended to the system message. The examples of the prompts are also in Table~\ref{tab:icl_methods}. We systematically varied all conditions, including retrieval based on note similarity, question similarity, question-type matching, and random selections. For the composed cases, we included scenarios with only correct examples, only incorrect examples, contrasted examples (correct and incorrect), or random selection. 

\begin{table*}[ht]
\centering
\small
\caption{Description of prompting strategies evaluated across models. All retrieval-augmented methods use GTR-T5-Base embeddings to retrieve the most similar discharge note from a pool of previously evaluated examples.}
\label{tab:icl_methods}
\begin{tabular}{llp{5.5cm}p{4.5cm}}
\toprule
\textbf{Condition} & \textbf{Group} & \textbf{Description} & \textbf{Prompt Keywords} \\
\midrule
Zero-shot & Baseline & Direct question answering with no examples or reasoning guidance. & \texttt{``You are a medical expert answering questions about discharge summaries.''} \\
\addlinespace
Positive ICL & RA-ICL & Retrieves a correctly answered example with the most similar discharge note and presents it as a demonstration. & \texttt{``Here is an example of a good answer: [Q]... [A]... Apply the same precision.''} \\
\addlinespace
Negative ICL  & RA-ICL & Retrieves an incorrectly answered example and shows both the mistake and the correction. & \texttt{``Here is a common mistake to avoid: [Incorrect]... [Correct]... Answer based only on what is explicitly stated.''} \\
\addlinespace
Contrastive ICL & RA-ICL & Combines a negative example (mistake + correction) followed by a positive example (good answer). & \texttt{``EXAMPLE OF A MISTAKE:... EXAMPLE OF A GOOD ANSWER:... Apply the same precision and avoid the same kinds of mistakes.''} \\
\addlinespace
CoT (evidence-first) & CoT & Instructs the model to first extract supporting evidence from the note, then derive the answer. & \texttt{``First, extract the specific evidence... Then provide your answer based solely on that evidence.''} \\
\addlinespace
CoT (conclusion-first) & CoT & Instructs the model to state the answer first, then explain its reasoning with reference to the note. & \texttt{``First state your answer, then explain your reasoning based on the discharge summary.''} \\
\midrule
\addlinespace
Self-Critique + RA-ICL & Med-HEAL & Model itself evaluates the zero-shot answer for clinical reasoning errors. Flagged answers are regenerated with a concise prompt that includes the critic's error diagnosis, and a retrieved correct example with the most similar discharge note. & \texttt{``A previous answer had this error: [diagnosis]. Provide a corrected answer in 2--3 sentences, based strictly on the discharge summary. Here is an example of a good answer: [Q]... [A]...''} \\
\bottomrule
\end{tabular}
\end{table*}

\section{Results}
\subsection{Generated Med-HEAL Dataset }
To construct the Med-HEAL dataset, we first evaluate BioMistral-7B in a zero-shot setting on 962 open-ended questions from EHRNoteQA. Each input consists of a patient discharge summary and a clinical question, and the model generates a free-text answer.

Answer correctness is determined through a two-stage evaluation pipeline.
\begin{enumerate}
\item GPT-4o acts as a judge that determines whether the generated answer conveys the same key clinical facts/conclusions as the clinician-verified ground truth given by EHRNoteQA, producing a binary correct/incorrect label.

\item GPT-4o analyzes incorrect answers to identify hallucination types, assigning scores on a 1–5 confidence scale for three categories: incomplete reasoning, factual errors, and fabricated sources, as discussed previously \cite{kim2025medical}. Scores of $\ge 4$ indicate that a hallucination type is likely present.
 
\end{enumerate}

The final dataset contains both the LLM-as-a-Judge and human reviewers' output and reasoning annotations, and we have two example cases in Table~\ref{tab:hallucination-examples}. In addition, Table~\ref{tab:biomistral_baseline} summarizes the results. BioMistral-7B achieves an overall accuracy of 63.8\%, with 348 incorrect answers, constituting the Med-HEAL dataset. Accuracy declines substantially as the number of discharge notes increases. When only a single note is provided, accuracy reaches 74.1\%, but drops to 62.3\% with two notes and 38.9\% with three notes. This trend suggests that the model struggles to synthesize information from longer and more complex clinical narratives.

We apply GPT-4o to categorize the incorrect answers into three hallucination types. Each answer might be labeled with multiple error types. 98.0\% incorrect answers are labeled with factual errors, followed by 82.2\% with incomplete reasoning, and 5.2\% with fabricated sources. These findings indicate that hallucinations in this task primarily arise from failures to integrate multiple pieces of evidence from patient records, rather than from inventing entirely unsupported facts. 


\begin{table*}[t]
\centering
\small
\caption{Examples of LLM hallucinations detected by both human expert and GPT-4o judge. BioMistral-7B open-ended answers on EHRNoteQA. Hallucinated content is \underline{underlined}.}
\label{tab:hallucination-examples}
\renewcommand{\arraystretch}{1.3}
\begin{tabularx}{\textwidth}{p{1.1cm} X}
\toprule
\multicolumn{2}{l}{\textbf{Case 1:} \textit{What was the cause of the patient's right subtrochanteric stress fracture, and how was it managed?}} \\
\midrule
\textbf{Ground Truth} & The patient developed the fracture due to the medication \textbf{phosphomax}, and the fracture was corrected by \textbf{total hip replacement}. \\[2pt]
\textbf{Incorrect Answer} & The cause of the patient's right subtrochanteric stress fracture was due to \underline{a fall from standing height while performing a squatting motion}. The fracture was managed with \underline{a right total femoral nail (TFN) surgery}. \\[2pt]
\textbf{Human Judge} & \texttimes~Incorrect --- Fabricated source (\checkmark), Factual error (\checkmark): The model fabricated the cause of the fracture (a fall) and the surgical procedure (TFN); the discharge summary attributes the fracture to the medication phosphomax and documents total hip replacement. \\[2pt]
\textbf{GPT-4o Judge} & \texttimes~Incorrect --- ``The medication is likely misidentified; the correct name is Fosamax (alendronate), which is associated with atypical femoral fractures, and the management typically involves surgical intervention such as intramedullary nailing, not specifically trochanteric fixation nailing.'' \\
\midrule
\multicolumn{2}{l}{\textbf{Case 2:} \textit{Did the patient's hemoglobin levels improve from 2166-11-22 to 2167-08-25?}} \\
\midrule
\textbf{Ground Truth} & The hemoglobin levels \textbf{improved} from 9.0 to 11.6. \\[2pt]
\textbf{Incorrect Answer} & \underline{No}, the patient's hemoglobin levels \underline{did not improve} from 9.0 g/dL on 2166-11-22 to 11.6 g/dL on 2167-08-25. \\[2pt]
\textbf{Human Judge} & \texttimes~~Incorrect --- Incomplete reasoning (\checkmark), Factual error (\checkmark): The model correctly states the values (9.0 $\rightarrow$ 11.6) yet contradicts itself by concluding they did not improve, directly misrepresenting the patient's clinical progress. \\[2pt]
\textbf{GPT-4o Judge} & \texttimes~~Incorrect --- ``The model falsely claims that the patient's hemoglobin levels did not improve, despite the data showing an increase from 9.0 to 11.6.'' \\
\bottomrule
\end{tabularx}
\end{table*}

\begin{table}[ht]
\centering
\small
\caption{BioMistral-7B zero-shot performance judged by GPT-4o on EHRNoteQA (962 questions). Hallucination types are among the 348 incorrect answers. (Confidence Level $\geq$ 4).}
\label{tab:biomistral_baseline}
\begin{tabular}{lcc}
\toprule
\textbf{Metric} & \textbf{Count} & \textbf{Rate} \\
\midrule
Total questions          & 962 & --- \\
Correct answers          & 614 & 63.8\% \\
Incorrect answers        & 348 & 36.2\% \\
\midrule
\multicolumn{3}{l}{\textit{Hallucination type (multi-labels, among 348 errors, classified by GPT-4o)}} \\
\quad Factual errors       & 341 / 348 & 98.0\% \\
\quad Incomplete reasoning & 286 / 348 & 82.2\% \\
\quad Fabricated sources   &  18 / 348 &  5.2\% \\
\bottomrule
\end{tabular}
\end{table}

\subsection{LLM Judgment Calibration}
To evaluate the reliability of our automated assessment, we benchmarked LLM-based judgments against independent human reviews from three medical students (A, B, and C). Reviewers A and B had completed their core clinical rotations, while Reviewer C was a first-year preclinical student. 

Initial inter-rater reliability among the three reviewers showed fair-to-moderate agreement ($\kappa = 0.287$). This was primarily due to Reviewer C’s systematically stricter labeling threshold, which lowered overall consistency. To mitigate this variance, we resolved conflicts via majority vote across a randomly sampled 100-item subset to establish a gold standard for comparison. 

Table~\ref{tab:openended_agreement} reports accuracy between each judge and the gold standard on a randomly sampled 100-item subset. Reviewers A, B, and C have 89.0\%, 87.0\%, and 80.0\% accuracy against the gold standard specifically. Meanwhile, instruction-tuned GPT-4o achieves 94.0\% agreement against the same gold standard. We ensure GPT-4o as an LLM Judge meet the criteria we obtained from the human gold standard in evaluation. 

Overall, these results suggest GPT-4o as a reliable primary evaluator for open-ended clinical QA on EHR data, and LLM-based evaluation can aligns well with human judgments, especially when we account for human inconsistency due to pre-bias or dual standards. In some cases provides more consistent labeling than individual human annotators. Consequently, we adopt a human-audited LLM evaluation pipeline, where GPT-4o performs the primary evaluation and human reviewers verify selected cases and annotations. Both the correctness labels and reasoning annotations produced through this process are included in the Med-HEAL dataset.

\begin{table}[ht]
\caption{Agreement for BioMistral-7B open-ended evaluation. Gold standard = 100 sampled questions agreed by human reviewers.}
\label{tab:openended_agreement}
\centering
\small
\begin{tabular}{lc}
\toprule
\textbf{Judge} & \textbf{Agreement} \\
\midrule
Reviewer A  & 89.0\% \\
Reviewer B  & 87.0\%  \\
Reviewer C & 80.0\% \\
GPT-4o    & 94.0\%  \\
\bottomrule
\end{tabular}
\end{table}

Since GPT-4o may act as a noisy judge, we apply a calibration approach suggested by the Rogan-Gladen estimator \cite{rogan1978estimating}, which corrects observed outcomes using the judge's performance on the gold-standard dataset. Specifically, we use 100 human-verified gold-standard examples to estimate a confusion matrix (Table \ref{tab:gpt-confusion-matrix}) for GPT-4o, from which we derive the true positive rate (TPR) and false positive rate (FPR). Assuming these error rates generalize to the full dataset, we calibrate the observed accuracy $\hat{p}$ given by GPT-4o
 to estimate the true accuracy $p$ as follows
\begin{equation} \label{eq:cali}
p = \frac{\hat{p} - \mathrm{FPR}}{\mathrm{TPR} - \mathrm{FPR}}.
\end{equation}
The calibrated results (Table \ref{tab:calibrated}) are more reliable and less biased, indicating true performance improvements between Med-HEAL and zero-shot across models.

\begin{table}[h]
\caption{Confusion matrix for GPT-4o judge}
\label{tab:gpt-confusion-matrix}
\centering
\begin{tabular}{c|cc|c}
\hline
\begin{tabular}{@{}c@{}}
\small GPT-4o Judge \\
\rule{2.5cm}{0.4pt} \\
\small Gold Standard
\end{tabular}
& 0 (wrong) & 1 (correct) & Total \\
\hline
0 & TN = 2 & FP = 2 & 13 \\
1 & FN = 4 & TP = 83 & 87 \\
\hline
Total & 15 & 85 & 100 \\
\hline
TPR (Sensitivity) & \multicolumn{3}{c}{$TP/(TP+FN) = 83/(83+4) = 0.954$} \\
FPR & \multicolumn{3}{c}{$FP/(FP+TN) = 2/(2+11) = 0.154$} \\
\hline
\end{tabular}
\end{table}

\begin{table}[ht]
\centering
\small
\caption{Calibrated Accuracy ($p$) of Zero-shot vs.\ best ICL performance across five models using the Med-HEAL pipeline Results.}
\label{tab:calibrated}
\begin{tabular}{llcccc}
\toprule
\textbf{Model} & \textbf{Zero-shot (\%)} & \textbf{Med-HEAL (\%)} & \textbf{$\Delta$ (\%)} & \textbf{$p$-value} \\
\midrule
BioMistral-7B      & $48.1 \pm 5.0$  & $50.4 \pm 3.3$ & $+2.3$ & $0.11$\\
DeepSeek-R1-8B      & $76.9 \pm 1.3$ & $85.9 \pm 0.9$ & $+9.0$ & $<0.001$ \\
Llama-3.1-8B       & $92.1 \pm 1.1$ & $94.1 \pm 1.6$ & $+2.0$ & $0.024$ \\
Qwen2.5-7B         & $91.6 \pm 1.7$ & $92.8 \pm 0.5$ & $+1.2$ & $0.02$ \\
Qwen3-8B           & $96.3 \pm 1.3$ & $98.0 \pm 1.7$ & $+1.7$ & $0.07$ \\
\bottomrule
\end{tabular}
\end{table}

\subsection{In-Context Learning for Mitigating Hallucination}
We evaluate whether the Med-HEAL dataset can mitigate hallucinations through retrieval-augmented in-context learning (ICL). Experiments are conducted on five open-source models commonly used in privacy-sensitive clinical environments: BioMistral-7B, DeepSeek-R1-Distill-8B, Qwen2.5-7B, Llama-3.1-8B, and Qwen3-8B. Since each summary generated by an LLM still needs a judge to determine whether it matches the ground-truth summary, we continue to use GPT-4o as the judge and report calibrated results using Eq.~\ref{eq:cali}. 

Table~\ref{tab:calibrated} shows the best calibrated result given by Med-HEAL, compared to the baseline zero-shot result. All five models have shown performance improvements over Zero-shot answers, ranging from 1\% to 9\%. We use McNemar's exact test to compare performance between zero-shot answers and observe significant improvements ($p < 0.05$) on DeepSeek, Llama3 and Qwen2.5. DeepSeek shows the most noticeable performance gain at 9\%, as its zero-shot answer accuracy is the lowest and its answers are easiest to correct. Qwen3 can convert a few incorrect answers to correct ones, but is limited by its near-ceiling effect ($96.3\%$); the improvement ($+1.7\%$) is very close to significant ($p=0.07$).

BioMistral is the base model, which has a minimum zero-shot performance ($48.1\%$), but does not benefit significantly from the self-critique pipeline for two reasons. First, BioMistral has a maximum context window of 4K tokens because its base model, Mistral-7B-Instruct-v0.1, can largely exhaust lengthy discharge notes, leaving insufficient capacity for additional instructions and hallucination examples. Second, the model has been fine-tuned on biomedical knowledge. BioMistral cannot follow instructions to point out its owner's mistake and frequently outputs unrelated BioMedical knowledge, likely learned during its fine-tuning. All indicate that BioMistral is unable to perform the self-critique task within the Med-HEAL pipeline.

\paragraph{\textbf{Effect of Med-HEAL Pipeline.}}
We compare several prompting strategies: (1) zero-shot baseline, (2) positive retrieval (correct example), (3)
negative retrieval (incorrect example with correction), (4) naive contrastive retrieval (incorrect + correct example), (5) evidence-first Chain-of-Thought (CoT), (6) conclusion-first (CoT), (7) Med-HEAL: Self-Critique and RA-ICL in the Table~\ref{tab:main_comparison} in the Appendix. 

Among all the strategies, we find that the simple retrieval methods, (2) (3) and (4), and the CoT methods, (5) and (6), do not work satisfactorily.

\section{Discussion}

This work introduces Med-HEAL, a framework for systematically collecting clinically realistic hallucination cases and mitigating them through a self-critique regeneration pipeline. Unlike prior approaches that rely on generic few-shot examples or simple retrieval of semantically similar cases \cite{feldman2023trapping}, Med-HEAL employs a multi-stage reasoning loop: the model first generates an answer, then critiques its own response to identify potential reasoning errors, and finally regenerates the answer when a potential hallucination is detected. This design enables models to correct reasoning failures through self-reflection without requiring parameter updates or additional training.

Our experiments across five open-source models on the EHRNoteQA benchmark (962 questions, 5-fold CV) reveal several critical patterns. First, naive retrieval-based ICL strategies provide little or no improvement over zero-shot performance and sometimes degrade accuracy. For example, BioMistral and Qwen2.5 performance drops under contrastive retrieval (in supplementary Table \ref{tab:main_comparison}. These results suggest that simply retrieving semantically similar demonstrations does not reliably guide models to avoid hallucinations in complex clinical reasoning tasks.

Second, Chain-of-Thought prompting produces model-dependent results. While models with stronger reasoning capabilities, such as DeepSeek, occasionally benefit from CoT prompts, the improvements are inconsistent across models and do not systematically reduce hallucinations.

In contrast, the Med-HEAL self-critique pipeline produces more consistent improvements across several models. Self-critique pipeline improves DeepSeek from 76.9\% to 85.9\% (+9.0\%), Qwen2.5 from 91.6\% to 92.8\% (+1.2\%), and Llama-3.1 from 92.1\% to 94.1\% (+2.0\%). The magnitude of the improvement also depends on the model's capability. Models with moderate baseline performance, such as DeepSeek and Qwen2.5, benefit most from self-critique, while the strongest model (Qwen3) shows limited improvement due to its already high baseline accuracy. This suggests that self-critique mechanisms are most effective when the model has sufficient reasoning ability to detect its own errors, yet still produces a nontrivial number of reasoning failures.

These results suggest that asking more capable models to explicitly examine and revise their own reasoning, with support from correct and hallucination examples, can effectively mitigate some hallucinations in EHR-based question answering.


Several limitations remain. First, the current Med-HEAL dataset is generated primarily using BioMistral-7B as the source model, which may limit the diversity of hallucination patterns captured. Expanding the dataset to include hallucinations generated by multiple models would improve coverage of possible failure modes. Meanwhile, we can fully investigate and explore the reliability of LLM-as-a-judge. Second, although the dataset includes rich annotations from both LLM judges and human reviewers, our current experiments mainly leverage correctness labels and example responses. Future work may explore alternative ways to utilize these annotations, including supervised fine-tuning, preference learning, or taxonomy-aware prompting strategies.

Finally, we observe that model accuracy decreases substantially as the number of discharge notes increases, suggesting that longer clinical contexts significantly increase task difficulty. Clinical narratives contain dense terminology, temporal dependencies, and heterogeneous information types, which may challenge current architectures when synthesizing information across long documents. Addressing this limitation will likely require improved context-handling strategies or architectures designed for long-context reasoning, or experiments on more capable models.

Overall, Med-HEAL demonstrates that hallucination-aware datasets, combined with self-critique prompting, provide a practical approach to improving the reliability of medical LLMs, particularly in tasks involving complex clinical narratives.

\section{Conclusion}

This paper presents Med-HEAL, a framework for studying and mitigating hallucinations in medical LLMs through clinically grounded evaluation and hallucination-aware prompting. Our experiments demonstrate that conventional retrieval-based in-context learning offers little benefit for this task, whereas a self-critique pipeline consistently improves performance across multiple open-source models. Combining self-critique with targeted demonstrations further improves accuracy by providing examples that address predicted reasoning failures. 

These findings suggest that enabling models to identify and correct their own reasoning errors is a promising direction for improving the reliability of medical LLMs. Med-HEAL provides both a dataset and a practical prompting strategy for advancing safer clinical AI systems.
 
\begin{acks}
We thank the Medical College of Wisconsin student evaluators who served as human judges of LLM-generated answers, with special thanks to Sara Saif and Caitlin Schwanke for major contributions to this effort: Michael Brianna, Gabriella Schultz, Teagan Blohowiak, Lauren Loftis, Evan Robinson, Ayman Alsalhi, and Sonu Bhandari. We also thank Kushali Darak from Concordia University Wisconsin for data collection support.
  
This work is partially supported by NSF CICI award 2517121.
\end{acks}

\bibliographystyle{ACM-Reference-Format}
\bibliography{paper-references}

\appendix

\section{Supplementary Materials}
The following supplementary materials provide extended background, dataset details, runtime analysis, and additional cross-model results.

\section{Related Work}\label{sec:related}
\subsection{Large Language Models in Medicine}
LLMs have been increasingly applied to clinical tasks such as diagnostic reasoning, patient question answering, and medical document analysis. Proprietary models including Med-PaLM \cite{singhalLargeLanguageModels2023}, Med-PaLM 2 \cite{singhal2025toward}, and other large commercial models demonstrate strong performance on medical benchmarks such as USMLE-style examinations \cite{jin2021disease}. General-purpose models such as GPT-4 have also shown strong medical reasoning capabilities despite not being exclusively trained for healthcare tasks \cite{achiam2023gpt,cabral2024clinical}.

To address privacy and deployment constraints in healthcare environments, a number of open-source medical LLMs have been developed by adapting general foundation models with biomedical corpora. Examples include Meditron \cite{chenMeditron70bScalingMedical2023}, BioMistral \cite{labrak2024biomistral}, PMC-LLaMA \cite{wu2024pmc}, and MedAlpaca \cite{han2023medalpaca}, which are trained or fine-tuned on resources such as PubMed \cite{jinPubmedqaDatasetBiomedical2019} and clinical guidelines. More recently, reasoning-oriented open models such as DeepSeek-R1 \cite{guo2025deepseek} and QwQ-32B \cite{yang2025qwen3} have shown promising performance in complex diagnostic reasoning tasks. These models are particularly attractive to healthcare institutions because they can run locally while maintaining competitive performance.

\subsection{Study of Medical LLM Hallucinations}
A growing body of work has developed benchmarks to evaluate large language models (LLMs) in medical settings, focusing on both clinical reasoning and hallucination detection. Many datasets are built from electronic health records (EHRs), medical literature, or standardized medical examinations. Among the most widely used sources are the MIMIC datasets—particularly MIMIC-III \cite{johnsonMIMICIIIFreelyAccessible2016} and MIMIC-IV \cite{johnson2020mimic}—which provide large collections of de-identified clinical records including discharge summaries, radiology reports, laboratory results, and treatment histories. These datasets enable the development of benchmarks that reflect realistic clinical workflows.

Several benchmarks derived from these resources evaluate LLM reasoning over clinical data. EHRNoteQA \cite{kweon2024ehrnoteqa} builds on MIMIC-IV and contains clinician-validated question–answer pairs requiring models to synthesize information across multiple patient notes, providing a realistic proxy for clinical decision-making tasks. Other EHR-based benchmarks such as CliBench \cite{ma2024clibench} and VeriFact-BHC \cite{chungVerifyingFactsPatient2025} similarly evaluate patient-specific reasoning and fact verification using real hospital records. Beyond EHRs, datasets such as MedCaseReasoning \cite{wu2025medcasereasoning} leverage thousands of clinical case reports from PubMed Central to evaluate diagnostic reasoning processes, while Med-HALT \cite{palMedhaltMedicalDomain2023} uses standardized medical examinations and PubMed abstracts to test hallucination robustness under adversarial prompts.

A number of datasets explicitly focus on medical hallucinations. MedHallu generates hallucinations through controlled LLM pipelines that introduce specific error types, such as evidence fabrication or misinterpretation of questions \cite{panditMedhalluComprehensiveBenchmark2025a}. MEDHALU constructs hallucinated responses based on a taxonomy of conflicts, including fact-conflicting and context-conflicting errors \cite{agarwal2024medhalu}. Med-HALT \cite{palMedhaltMedicalDomain2023}  evaluates hallucinations through adversarial testing, such as presenting incorrect answers or nonsensical medical questions \cite{palMedhaltMedicalDomain2023}. These benchmarks provide important testbeds for evaluating model reliability, though many primarily focus on evaluation rather than generating clinically realistic hallucination cases for training or mitigation.

Reliable evaluation of medical LLM outputs remains challenging. Human expert evaluation remains the gold standard because clinicians are better able to identify subtle medical inaccuracies and contextual errors \cite{kim2025medical}. However, human annotation is expensive and difficult to scale.
To address this limitation, recent work has explored LLM-as-a-Judge approaches, where large models are used to evaluate or rank model outputs like relevance, coherence, accuracy, helpfulness, or adherence to instructions \cite{guSurveyLlmasajudge2024,zheng2023judging,liu2023g}. This paradigm has been widely adopted in general NLP benchmarks such as MT-Bench and AlpacaEval, and is increasingly used in medical datasets. For example, ClinBench-HPB \cite{li2025clinbench} employs a dual-LLM evaluation framework, while datasets such as MedCaseReasoning, VeriFact, and MedHallu use LLMs to identify hallucinated statements or evaluate reasoning quality. Current research often combines automated LLM evaluation with selective human verification in human-in-the-loop workflows.

\subsection{EHRNoteQA}
EHRNoteQA is not focused on a single medical specialty; instead, it covers a broad range of real-world clinical scenarios by categorizing queries into 10 diverse medical topics\cite{kweon2024ehrnoteqa}. Because the benchmark is built from MIMIC-IV discharge summaries encompassing complex hospital admissions, it reflects comprehensive, multi-faceted patient care. The questions in the benchmark are designed to mirror the wide variety of inquiries a physician might actually make when reviewing patient records. 

\begin{table}[ht]
\centering
\small
\caption{EHRNoteQA Question Topic Coverage}
\label{tab:ehrnoteqa-topics}
\begin{tabular}{lp{5.5cm}r}
\toprule
\textbf{Category} & \textbf{Example} & \textbf{\%} \\
\midrule
Treatment      & What was the treatment provided for the patient's left breast cellulitis?                          & 64\% \\
Etiology       & Why did the patient's creatinine level rise significantly upon admission?                          & 20\% \\
Assessment     & Was the Mitral valve repair carried out successfully?                                              & 19\% \\
Problem        & What was the main problem of the patient?                                                          & 19\% \\
Test Results   & What were the abnormalities observed in the patient's CT scans?                                    & 14\% \\
Sign/Symptom   & What was the presenting symptom of the patient's myocardial infarction?                            & 12\% \\
History        & Has the patient experienced any surgical interventions prior to the acute appendicitis?             & 12\% \\
Plan           & What is the future course of action planned for patient's left subclavian stenosis?                 &  5\% \\
Vitals         & What was the range of the patient's blood pressure during second stay?                             &  3\% \\
Instruction    & How was the patient instructed on weight-bearing after his knee replacement?                       &  3\% \\
\bottomrule
\end{tabular}
\end{table}

\subsection{In-Context Learning}
In-context learning (ICL) enables LLMs to perform tasks by conditioning on a small set of demonstration examples without updating model parameters \cite{garg2022can,akyurek2022learning}. Prior studies show that retrieving demonstrations which are semantically similar to the query significantly improve performance and reduce hallucinations in general domains \cite{luo2024dr}. Retrieval-based ICL methods often rely on lexical or semantic retrievers such as BM25 \cite{robertson2009probabilistic} or other embedding-based similarity \cite{rubin2022learning}.

Several prompt-design strategies have been proposed to further improve reliability. Chain-of-Thought (CoT) \cite{wei2022chain} prompting encourages models to generate explicit reasoning steps, reducing logical errors and incomplete reasoning. Structured reasoning frameworks, such as Chain-of-Medical-Thought \cite{jiang2025comt}, guide models through clinically meaningful reasoning stages and have been shown to reduce severe hallucinations. Recent studies in other domains  \cite{wei2022chain,dhuliawala2024chain,sun2025improving} also suggest that providing negative examples, corrections, or expert feedback in prompts can improve the model’s ability to detect and avoid hallucinations.

Despite these advances, most existing work focuses on prompting strategies or evaluation benchmarks. Relatively little research explores how clinically realistic hallucination examples themselves can be systematically constructed and leveraged within contrastive in-context learning frameworks to directly train models to recognize and avoid hallucinated reasoning. Addressing this gap motivates the framework proposed in this work.

\subsection{Per-item Runtime Cost Analysis}
\label{sec:supp_runtime}

Table~\ref{tab:runtime} summarizes the average runtime per item for the zero-shot baseline compared to the Med-HEAL pipeline. Measurements were conducted on a local workstation (Intel i9-13900K, RTX 4090).

\begin{table*}[t]
    \centering
    \caption{Per-item runtime comparison between the zero-shot baseline and the Med-HEAL pipeline.}
    \small
    \begin{tabular}{lccc}
        \toprule
        \textbf{Model} & \textbf{Zero-shot Baseline} & \textbf{Med-HEAL Pipeline} & \textbf{Multiplier} \\
        \midrule
        \textit{Local Models} & & & \\
        Qwen2.5       & $\sim$5 sec      & $\sim$17 sec  & $\mathbf{\sim3.4\times}$ \\
        Llama-3.1     & $\sim$5 sec      & $\sim$16 sec  & $\sim$3.2$\times$ \\
        BioMistral    & $\sim$10 sec     & $\sim$30 sec  & $\sim$3$\times$ \\
        DeepSeek-R1   & $\sim$10--15 sec & $\sim$30 sec  & $\sim$2--3$\times$ \\
        Qwen3-8B      & $\sim$10--15 sec & $\sim$38 sec  & $\sim$2.5--4$\times$ \\
        \midrule
        \textit{Commercial Cloud API} & & & \\
        GPT-4o        & $\sim$4 sec      & $\sim$12 sec  & $\sim$3$\times$ \\
        Claude-3.5    & $\sim$5 sec      & $\sim$15 sec  & $\sim$3$\times$ \\
        \bottomrule
    \end{tabular}
    \label{tab:runtime}
\end{table*}

\subsection{Cross-model Comparison Benchmark on EHRNoteQA}

\paragraph{\textbf{Baseline Performance.}}
Zero-shot performance varies across models, with BioMistral achieving 53.8\% accuracy, while larger general-purpose models such as Qwen3 and Llama-3.1 reach over 89\%. These results highlight substantial variation in baseline reasoning capability across models when answering clinical questions from discharge summaries.

\paragraph{\textbf{Effect of Retrieval-Based ICL.}}
Across all models, standard retrieval-augmented ICL provides little or no improvement over the zero-shot baseline. In some cases, performance decreases slightly when positive, negative, or contrastive demonstrations are added. For example, BioMistral drops from 53.8\% to 48.4\% under contrastive ICL, and Qwen2.5 decreases from 88.7\% to 86.9\%. These results suggest that simply retrieving semantically similar examples does not effectively guide models to avoid hallucinations in complex clinical reasoning tasks.

\paragraph{\textbf{Effect of Chain-of-Thought Prompting.}}
Chain-of-Thought prompting produces mixed results across models. Models with stronger reasoning capabilities benefit modestly from CoT prompts. For instance, DeepSeek improves from 76.9\% to 78.3\% under conclusion-first CoT, while Qwen3 achieves its highest accuracy of 94.3\% with the same prompting strategy. However, CoT does not consistently improve performance across all models, and in some cases slightly reduces accuracy.

\paragraph{\textbf{Summary of Findings.}}
Overall, the results indicate that naive retrieval-based ICL provides limited benefit for clinical question answering tasks. In contrast, hallucination-aware prompting strategies that incorporate a self-critique step and targeted demonstrations can produce modest but consistent improvements across several models. These findings highlight the importance of error-aware prompting mechanisms when applying in-context learning to mitigate hallucinations in medical LLMs.

\begin{table*}[t]
\centering
\caption{Cross-model comparison of ICL strategies on EHRNoteQA (5-fold CV, $N{=}962$, GPT-4o evaluation). For self-critique conditions, non-critiqued samples retain their zero-shot answers. Best result per model in \textbf{bold}.}
\label{tab:main_comparison}
\small
\begin{tabular}{llccccc}
\toprule
Group & Condition & BioMistral & DeepSeek & Llama-3.1 & Qwen2.5 & Qwen3 \\
\midrule
\multirow{1}{*}{Baseline} & Zero-shot & 53.8$\pm$4.1 & 76.9$\pm$1.0 & 89.1$\pm$0.9 & 88.7$\pm$1.4 & 92.4$\pm$1.1 \\
\midrule
\multirow{3}{*}{RA-ICL} & Positive ICL & 52.2$\pm$3.8 & 74.8$\pm$1.4 & 89.0$\pm$2.3 & 88.6$\pm$1.3 & 93.1$\pm$0.9 \\
  & Negative ICL & 49.7$\pm$3.2 & 75.9$\pm$0.6 & 89.0$\pm$1.0 & 86.9$\pm$1.2 & 92.0$\pm$1.6 \\
  & Contrastive ICL  & 48.4$\pm$2.9 & 74.7$\pm$3.2 & 89.0$\pm$1.1 & 86.9$\pm$1.4 & 93.0$\pm$0.5 \\
\midrule
\multirow{3}{*}{CoT} & CoT (evidence-first) & 52.9$\pm$2.7 & 76.3$\pm$1.8 & 86.5$\pm$0.7 & 87.7$\pm$1.8 & 93.7$\pm$0.4 \\
  & CoT (conclusion-first) & 52.7$\pm$4.4 & 78.3$\pm$1.9 & 89.1$\pm$2.2 & 87.4$\pm$2.2 & 94.3$\pm$1.0 \\
\midrule
\multirow{1}{*}{Med-HEAL} & \textbf{Self-Critique+RA-ICL} & 50.8$\pm$1.8 & \textbf{81.9$\pm$2.6} & \textbf{90.1$\pm$1.4} & \textbf{92.6$\pm$1.6} & 93.4$\pm$1.5 \\
\bottomrule
\end{tabular}
\end{table*}

\end{document}